\documentclass[letterpaper]{article}

\usepackage{natbib,alifeconf}  
\usepackage{amsmath}
\usepackage[hidelinks]{hyperref} 
\usepackage{url,cleveref}
\usepackage{booktabs}

\usepackage{dsfont}

\usepackage{lipsum}

\makeatletter
\renewcommand{\paragraph}{%
  \@startsection{paragraph}{4}%
  {\z@}{1.75ex \@plus 1ex \@minus .2ex}{-1em}%
  {\normalfont\normalsize\bfseries}%
}
\makeatother

\newcommand\blfootnote[1]{%
  \begingroup
  \renewcommand\thefootnote{}\footnote{#1}%
  \addtocounter{footnote}{-1}%
  \endgroup
}

\title{Stability and Geometry of Attractors in Neural Cellular Automata}

\author{
    Mia-Katrin Kvalsund$^{1}$
    \and
    James Stovold$^2$ \\
    \mbox{}\\
    $^1$Department of Physics, University of Oslo, Norway \\
    $^2$Department of Computer Science, University of York, UK \\
    \url{m.k.o.kvalsund@fys.uio.no}
} 

\begin{document}

\maketitle

\begin{abstract}
\vspace{-0.5em}
    Throughout the literature on Neural Cellular Automata (NCAs), it is often taken for granted that the systems learn attractors. This is shown through evolving the system for many timesteps and noting visual similarity to the goal state. There remain many questions after such an analysis. Namely, what kind of attractors do we have? Is their behavior ordered or chaotic? Can we estimate stability over very long time horizons? What really happens in the attractor when perturbations are applied? In this paper, we present a case study to help answer these questions, with methods drawn from the literature on dynamical systems theory. 
    
    We use the growing gecko NCA of \cite{mordvintsev2020growing} with deterministic cell updates as a case study. To the best of the authors' knowledge, we present the first visualizations of NCA attractor dynamics. We also analyze them using the Lyapunov and Fourier spectra, to reveal that the NCA displays oscillatory, periodic and quasi-periodic behavior, and that these behaviors arise early during training. This challenges the belief that NCAs learn fixed point attractors. Finally, we show that large perturbations to the attractor states can throw the NCAs into a secondary mode separate from the original attractor. We hope that this initial foray into NCA attractor dynamics expands the toolkit for NCA researchers to analyze the robustness and stability of their systems. 
\end{abstract}

\vspace{-0.5em}
\noindent{}Submission type: \textbf{Full Paper}\\
\noindent{}Data/Code available at: \url{https://github.com/mia-katrin/attractors}
\blfootnote{\textcopyright  2026 Kvalsund et al. Published under a Creative Commons Attribution 4.0 International (CC BY 4.0) license.}

\vspace{-0.5em}
\section{Introduction}

NCAs are an ideal model for studying self-organizing behavior defined from a macroscopic perspective. The ability to train behavior into a distributed substrate is very powerful, with applications ranging from the study of preserved genetic codes in evolution~\citep{chow_developmentnecessitatesevolutionarily} and models of the neocortex~\citep{kvalsund_sensormovementdrives} to biology-driven artificial intelligence~\citep{hartl_neuralcellularautomata} and the basis for computational substrates~\citep{pontesfilho_reservoircomputingevolved, barandiaran_growingreservoirsdevelopmental}.

As NCAs become used in increasingly more critical applications, such as cell microscopy image classification \citep{yang2025hierarchical, yang2025attention} and medical segmentation \citep{kalkhof2025med}, it becomes ever more important to understand NCAs' behavior. This is the philosophy of explainable AI \citep{dwivedi2023explainable}. Due to the black-box nature of neural networks, however, we are not yet able to fully understand what the model has learned. There have been a few attempts at improving stability empirically~\citep{stovold_identityincreasesstability, stovold_emergentmovementincreased}, and at interpreting the behavior through the lens of partial differential equations (PDEs) such as \citep{niklasson_asynchronicityneuralcellular, mordvintsev_differentiableprogrammingreaction, pajouheshgar_discoveringpartialdifferential}, but the complexity of any derived equations makes analytical techniques intractable. Motivated by this, a recent paper analyzes attractor landscapes through dimensionality reduction and topological analysis \citep{stovold2026landscapes}. Following this work, we take an alternative approach in applying techniques from dynamical systems theory to analyze the temporal dynamics of the learned behavior.

Dynamical systems theory can be a powerful tool for understanding NCA attractors. NCAs are dynamical systems, just like recurrent neural networks (RNNs), where the time series $x_{i+1} = f(x_i)$ depends on the substrate state $x_i$ and the update function $f$. NCAs are typically discrete but can also display continuous dynamics \citep{pajouheshgar_discoveringpartialdifferential}. Unlike a general RNN, the NCA is often meaningfully trained to have global attractor states through pool-training and perturbations \citep{mordvintsev2020growing}.

NCAs are commonly assumed to have attractors \citep{mordvintsev2020growing, randazzo2020self, hartl_neuralcellularautomata} that may be fixed point attractors (as in \citet{tesfaldet2022attention}). For the self-classification of ``chimeric'' digits of \citet{randazzo2020self}, the system was visually determined to form oscillating attractors when struggling to classify. In the work of \citet{grattarola2021learning}, the presence of oscillatory attractors was determined from the loss graph oscillating. In both these papers, oscillations were associated with a failure mode of the model, and it was assumed that the attractor would converge to a fixed point when functioning correctly. Knowing the exact type of attractor can be important, as it helps determine the long-term behavior and stability of the system. 

In this paper, we use the growing gecko system of \citet{mordvintsev2020growing} as a case study to explore attractor visualization and classification. To the best of our knowledge, we present the first visualized NCA attractors. We analyze the attractors' stability and type using numerical tools that are not affected by dimensionality reduction tools. Further, we determine that this system displays oscillatory dynamics in all converged states that were previously thought to be fixed. Finally, we show what robustness to perturbations looks like at the attractor. 

\section{Theory}

\begin{figure*}
    \centering
    \includegraphics[width=\linewidth]{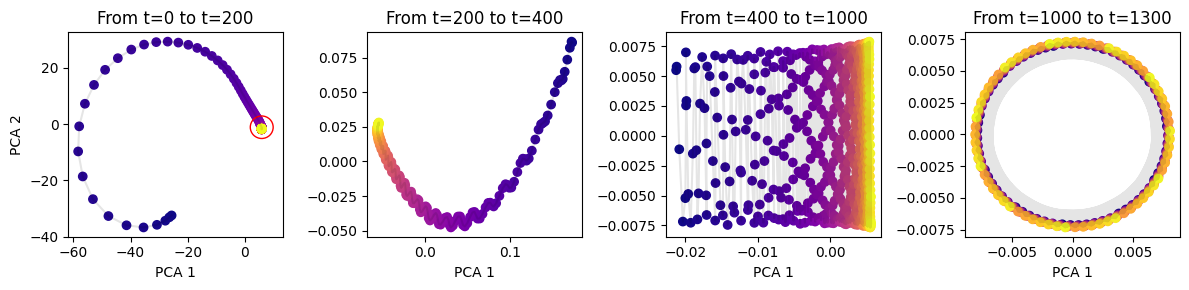}
    \caption{The burn-in of the attractor. From timestep $t$=0 to 200, the system moves a large distance. At timestep 60 (red circle), the system has already converged to the goal state. From timestep 200 to 400, the system is still moving (though the distances are small, notice the axes). From timestep 400 to 1000, the system finally converges, and from 1000 onward, the system no longer moves meaningfully along the rotational axis. Time intervals are transformed by separate PCA transforms.}
    \label{fig:burn_in}
\end{figure*}

\subsection{Neural Cellular Automata}

Neural Cellular Automata (NCAs) are a neural network generalization of the cellular automaton. Given a computational substrate $S$, a rule $f$ is applied locally in a neighborhood $N(x,y)$ at position $(x,y)$ in the substrate. The substrate is updated as such
\begin{equation}
    S(x,y) \leftarrow S(x,y) + f(N(x,y)) 
\end{equation}
for all positions $(x,y)$. However, many NCAs employ a stochastic update, such that 
\begin{equation}
    S(x,y) \leftarrow \begin{cases} 
      S(x,y) + f(N(x,y)) & \text{if } p < P \\
      S(x,y) & \text{otherwise} 
   \end{cases}
\end{equation}
where $p$ is sampled from a uniform distribution and P is the update rate, typically 50\%. The original justification for this was for biological accuracy, however it very likely also introduces further robustness in the system~\citep{niklasson_asynchronicityneuralcellular}. Including stochastic updates means that the system is a random dynamical system, which is a more complicated subtype. We have therefore opted not to use stochastic updates, focusing instead on the simpler deterministic case for this initial exploration of NCA attractors. 

NCAs always operate across time, and as such they are a form of recurrent neural network. Typically, the substrate acts as the system's recurrence, and only a select few channels are penalized by loss---the rest are left for memory and computation. This leaves a large subspace where the hidden channels are allowed to vary, meaning that there are infinitely many possible trajectories and attractors within this subspace. A well-performing NCA may learn one or multiple of these. 

NCAs are often trained with a sample pool. This means that a batch will consist of some percentage of the seed state (the initial state for the system to train from) and the rest of the batch will be previously achieved substrate states. Effectively, the system will learn to converge to the goal state for a wide array of starting conditions, not just the seed state. This means that the system will learn a more robust convergence than a fragile seed-to-attractor path.  

The dimensionality of an NCA is typically dependent on the problem. In the case of the growing gecko in \citep{mordvintsev2020growing}, the NCA is trained to maintain the image of a green gecko. This image is 40$\times$40. Further, it has 16 channels in the substrate. The system is also trained with a padded substrate (to avoid edge effects for the growing image), which adds another 16 pixels all around. As such, without padding, the whole substrate is a 40$\times$40$\times$16 $=$ 25600-dimensional flat vector. With padding, it is 72$\times$72$\times$16 $=$ 82944-dimensional.

\subsection{Attractors}

An attractor in the context of dynamical systems is defined as a set of points or trajectories that all trajectories nearby will approach as $t \rightarrow \infty$. It is said to be globally attracting if all trajectories approach the attractor (\cite[p.\ 130]{strogatz2001nonlinear}; \cite[p.\ 13]{datseris2022nonlinear}). 

There are many types of attractors:

\textbf{Fixed point:} In general, a fixed point is defined by $f(x*) = x*$ \citep[p. 4]{datseris2022nonlinear}. In other words, all points on the attractor are $x*$.

\textbf{Limit cycle/periodic:} A limit cycle is defined as $x_{t+T} = x_t$ where $T$ is finite \citep[p.\ 4]{datseris2022nonlinear}. In other words, states on the attractor repeat within a trajectory.  

\textbf{Quasi-periodic:} Although oscillatory like the limit cycle, a quasi-periodic trajectory never closes. There exists no number $T$ such that $x_{t+T} = x_t$. However, each variable in $x$ can be described as a function  of $k$ angles, and each trajectory progresses on a $k$-torus \citep[p.\ 30]{datseris2022nonlinear}. In simpler terms, quasi-periodic motion can be quite complex, but always oscillatory. 

\textbf{Chaotic, strange attractor:} Non-periodic and very sensitive to initial conditions, with fractal geometry. Strange attractors are difficult to predict long-term \citep[p.\ 40]{datseris2022nonlinear} \citep[p.\ 333]{strogatz2001nonlinear}.   

\subsection{Lyapunov spectrum}

For a dynamical system, the rate of change between two very close points can be measured by the highest Lyapunov exponent ($\lambda_1$). This is because, for a very small distance $\delta_0$ between two points at time $0$, the distance at time $t$ is:
\begin{equation}
    \delta_t \approx \delta_0 e^{\lambda_1 t}
    \label{eq:distance}
\end{equation}
As time goes to infinity, the largest rate of change dominates in the system \citep[p.\ 38]{datseris2022nonlinear}. A negative exponent means the system can be seen as Lyapunov stable, meaning perturbations will shrink and decay. If it is positive, the system has chaotic dynamics, meaning perturbations grow and the system is sensitive to initial conditions.

\enlargethispage*{2em}
For a D-dimensional system, you will have an ordered set of Lyapunov exponents
$$\lambda_1 \geq \lambda_2 \geq ... \geq \lambda_D$$
which is the Lyapunov spectrum \citep[p.\ 41]{datseris2022nonlinear}. Estimating the whole spectrum, or the upper $n$ exponents, can give you additional information about the type of attractor. In table \ref{tab:spectrum}, we note some characteristic spectra for attractors. Note that only non-linear systems can be chaotic \citep[p.\ 44]{datseris2022nonlinear}.

\begin{table}[ht]
    \centering
    \begin{tabular}{c|r}
        Attractor type & Lyapunov Spectrum \\
        \hline
        Fixed point & $- \cdots -$ \\
        Limit cycle/periodic & $0 - \cdots -$ \\
        k-torus/quasi-periodic & $\underbrace{0 \cdots 0}_{k \geq 1} - \cdots -$ \\
        Strange attractor & $\underbrace{+ \cdots +}_{l \geq 1} \underbrace{0 \cdots 0}_{k \geq 1} - \cdots -$ \\
    \end{tabular}
    \caption{Attractors and their characteristic spectrum. Strange attractors are a wider topic, and depending on how high the number $l$ and $k$ gets, there are more classifications of strange attractors within the cateory. Adapted from \citep{sandri1996numerical}.}
    \label{tab:spectrum}
\end{table}

\begin{figure}[ht]
    \centering
    \includegraphics[width=0.7\linewidth]{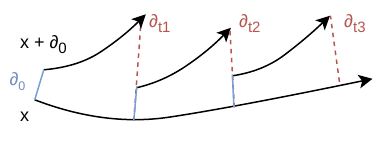}
    \caption{Estimating rate of change using finite differences. Adapted from \citep[p. 39]{datseris2022nonlinear}}
    \label{fig:distance_lyap}
    \vspace{-1em}
\end{figure}

The highest exponent can be calculated easily by evolving two different trajectories, $x$ and $x + \delta_0$, where $\delta_0$ is some random vector scaled by a small number $\epsilon$. The exponent can be approximated by 
\begin{equation}
    \lambda_1 \approx \frac{1}{N} \sum_{t = 0}^N \ln{\bigg(\frac{\delta_t}{\delta_0}\bigg)}
    \label{eq:finite_difference}
\end{equation}
For every step, a scaling of the vector $\delta_t = f(x + \delta_0) - f(x) $ back to the same norm as $\delta_0$ should be included to measure rate of change along the trajectory (see fig.\ \ref{fig:distance_lyap}). To estimate the whole spectrum, start instead with a reference point $x$ and $n \leq D$ orthonormal vectors. For each vector, evolve through $f$ and estimate the respective Lyapunov exponents after equation \ref{eq:finite_difference}. After each timestep, re-orthonormalize the vectors \citep[p. 211]{datseris2022nonlinear}. 

This method is relatively fast and gives an answer to whether the system is chaotic but is sensitive to the $\epsilon$ chosen. To avoid this problem, a Jacobian-vector product can be calculated directly. However, this method is slower and does not noticeably increase accuracy.    

The Van der Pol attractor \citep{van1926lxxxviii} and the famous Lorenz attractor \citep{lorenz1972predictability} are shown in fig.\ \ref{fig:fourier_explainer_figure}. As an example of Lyapunov spectra, the spectrum associated with Van der Pol is $\lambda_1=0$, indicating periodic motion and $\lambda_1, \lambda_2$ are weakly negative values. For the Lorenz attractor, we know that the spectrum is $\lambda_1 = 0.906$, $\lambda_2 = 0$, and $\lambda_3 = -14.573$. $\lambda_1$ indicates chaotic dynamics, while $\lambda_3$ indicates a strongly contracting direction. Our estimates for Lorenz' spectrum is 1.005, 0.003, and -14.953, showing that numerical estimates are often imperfect (as in \cite{sandri1996numerical}). The precision is dependent on floating point precision and the finite time chosen.   

\subsection{Basins of attraction}

The basin of an attractor is another important way to view stability. It is defined for an attractor set $A$ as the set of all initial conditions $x_0$ that satisfy the limit $||x_t - A|| \rightarrow 0$ as $t \rightarrow \infty$ \cite[p. 13]{datseris2022nonlinear}. In other words, a basin is defined by every starting condition that falls into the attractor. As such, the basin is an important consideration of sensitivity to perturbations: a small basin makes for a fragile attractor, while a large basin makes for a very robust attractor. An attractor can be Lyapunov stable but not robust to most perturbations because the basin is small.  

\subsection{Fourier spectrum}

A discrete time series can be represented as a linear combination of amplitudes multiplied by increasing frequencies of sines and cosines. As such, we can transform a time series of length $N$ into an array of amplitudes of length $(N//2+1)$. Taking the squared absolute value gives us the power spectrum of the time series. In short, we can determine with which frequencies a time series oscillates by inspecting the peaks in the spectrum. For attractors, we may average the spectra across all variables to get the global system's frequencies \citep[pp. 30-32]{datseris2022nonlinear}.

\begin{figure}[h!]
    \centering
    \includegraphics[width=0.9\linewidth]{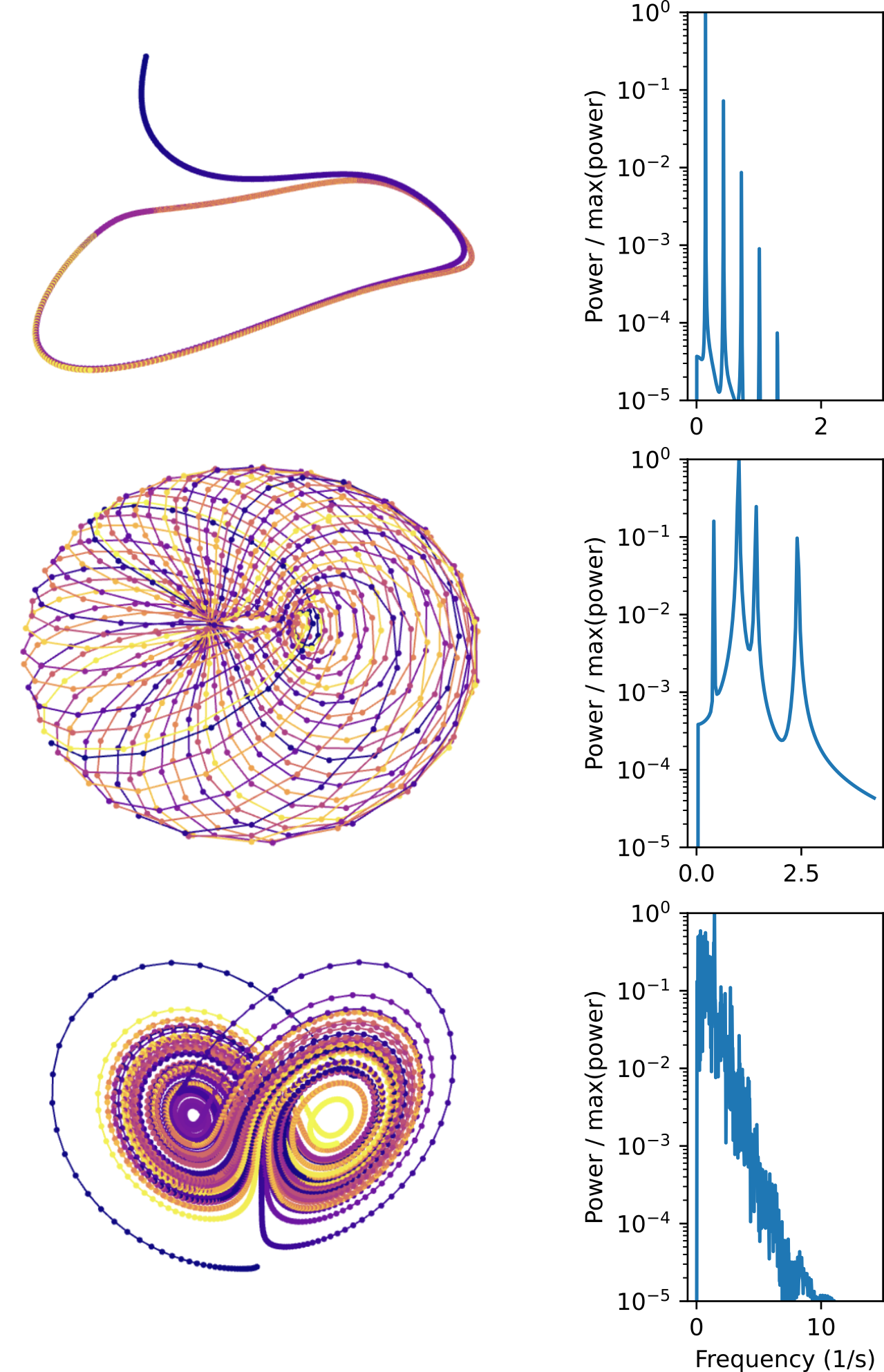}
    \caption{The Fourier spectra (right) of three attractors (left). From top to bottom: Van der Pol (periodic), a torus (quasi-periodic), and Lorenz (chaotic).}
    \label{fig:fourier_explainer_figure}
    \vspace{-0.5em}
\end{figure}

For periodic motion, we can expect to find one peak at the base frequency ($f$), plus harmonics ($m*f, m \in \mathds{N}$, where $\mathds{N}$ is all positive integers). In fig.\ \ref{fig:fourier_explainer_figure}, we get 5 clear peaks: 0.144, 0.432, 0.720, 0.856, 1.008. We observe that the base frequency is $f$=0.144, and that 0.432=3$f$, 0.720=5$f$, 0.856$\approx$6$f$, and 1.008=7$f$. Not all harmonics are necessarily present or equally strong. Note also that some harmonics are approximate: Because of rounding and sampling effects, we need to choose a tolerance for considering something a harmonic or linear combination. 

For k-th order quasi-periodicity, we can expect to find more than k sharp peaks. That is, linear combinations of the k-frequencies will also be present ($m_1 f_1  + m_2 f_2$, $m_1, m_2\in \mathds{Z}$, where $\mathds{Z}$ is all negative and positive integers, including 0). In fig.\ \ref{fig:fourier_explainer_figure}, we get 4 clear peaks: 0.4125, 1.0125, 1.425, and 2.4. We observe that the base frequencies are $f_1$=0.4125 and $f_2$=1.0125: The other two peaks are linear combinations: 1.425=$f_1+f_2$ and 2.4$\approx1f_1+2f_2$. A further requirement is that $f_1/f_2$ should be an irrational number---because of discreteness and rounding, this cannot easily be shown on a computer. 

Power spectra of chaotic attractors look noisy, showing a broadband spectrum. Although sometimes remnants of peaks are present, showing underlying, unstable periodic structure \citep[p. 31]{datseris2022nonlinear}.   

\subsection{Principal Component Analysis}

Principal Component Analysis (PCA) is a method of dimensionality reduction. It finds the direction of greatest variance, and all subsequent orthogonal vectors that are in the direction of greatest variance and uses this as a new basis for the data \citep[p.\ 26]{vidal2016principal}. PCA is a linear mapping so there is a risk of gluing separate structures together along axes of variation (e.g.\ an empty sphere projected from 3D to 2D will not have the empty room inside preserved as the top and bottom are fused together).

After fitting a PCA transform, we can extract the cumulative ratio for explained variance. As we add components, this ratio will go towards 100\%. However, if we choose a threshold $\tau$ to stop adding components, we can reason about the intrinsic dimensionality of our data. If $d$ components explain $\tau$\% of the variance, the data is commonly assumed to be $d$-dimensional. The intrinsic dimensionality and effective dimensionality may still differ, as PCA is a linear transform and may over- or underestimate dimensionality of nonlinear objects or noisy objects \citep{del2021effective}.  

PCA is very sensitive to scaling. Therefore, data is often scaled before PCA using standard scaling, which centers and scales each axis by its standard deviation. 

\section{Methods}

\begin{figure*}
    \centering
    \includegraphics[width=0.8\linewidth]{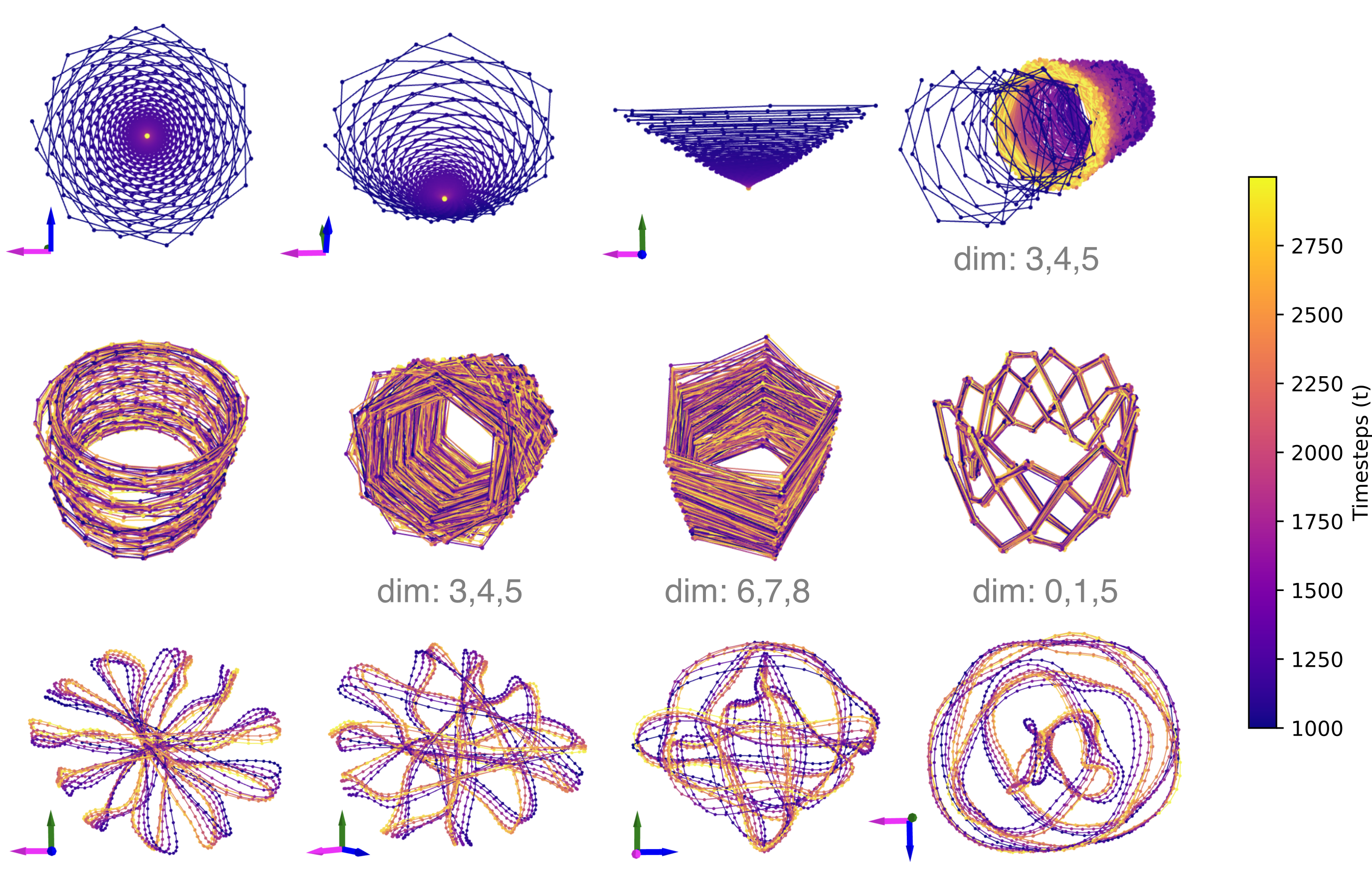}
    \caption{The attractors of the green gecko. Top: Version 1. Middle: Version 2. Bottom: Version 3. Unless otherwise specified, PCA axis 0 to 2 are used. Axes are only shown where relevant. Magenta, blue, and green are the first, second, and third most dominant PCA-axes respectively.}
    \label{fig:attractor_figure} 
\end{figure*}

\paragraph{Attractor data collection}

The system used was the NCA from \cite{mordvintsev2020growing}. The same code was used, except for the cell update probability, which was set to 100\%. Importantly, the system was trained with pool-training and perturbations. 

We trained three models: version 1, 2, and 3, that differ only by their random seed initialization. All models solve the task (i.e. maintain the shape of a green gecko over time and under large perturbations). 

To record from an attractor, it is important to include a ``burn-in'' phase, where the NCA is allowed to converge to the attractor from the seed state. This burn-in phase can be of variable length depending on the system: in these experiments, around 2000-4000 timesteps were used, depending on the attractor (see fig.\ \ref{fig:burn_in}).

After the burn-in, the system is assumed to be on the attractor. From there, the Lyapunov spectrum can be estimated. Alternatively, the system can be run for another thousand steps, after which a Fourier and PCA analysis can be performed, and the attractor can be plotted. 

\paragraph{Scaling}

We scale the data before performing PCA using sci-kit learn's StandardScaler. However, for version 1 of the NCA, we were fortunate enough to get a very simple attractor that can be visualized without scaling. Therefore, we have not scaled version 1 in some figures, so that we can show the magnitude of the attractor through the axes' values. We specify where scaling is not done. On the other attractors, the axes' values were removed because they are not informative.  

\paragraph{Perturbation}

To visualize convergence to the basins of attraction, we applied perturbations to a state on the attractor and let the NCA evolve from this until it reached the attractor (1500 steps). For large perturbations, we used the circle masks that the NCA has been trained to recover from. Small perturbations consisted of noise sampled from a normal distribution with mean 0 and standard deviation 0.002 (the scale of the attractors is very small). The noise was multiplied with the Boolean ``living mask'' and then applied to the attractor state. This ``living mask'' is from the original work and is a way of only updating the active cells that are forming the gecko (as opposed to the whole substrate).  

After collecting the perturbed trajectories of length 1500, they were scaled by a scaler and reduced by a PCA transform to be visualized. The large perturbations were visualized with the original attractor's scaler and PCA transform. For the small perturbations, both transforms were fitted anew: The original attractor and new attractor trajectories were combined into one vector, after which the transforms were fitted on that vector. The perturbed trajectories were assumed to be converged after 1000 steps, and so the last 500 steps were used to fit the transforms. Lastly, the new attractors found from the perturbed trajectories were visualized with their own fitted transforms. 

\section{Results}

\paragraph{Visualization} 


In fig.\ \ref{fig:attractor_figure}, the scaled dynamics of each attractor is shown. We can immediately conclude that none of these are fixed points. The PCA transformations are markedly different from each other: version 1 is dominated by slow dynamics, spiraling even after 1000 timesteps, and version 3 is a much more complex geometry than the other two. We will show with the further analysis that these visual differences are substantiated, and not just an artifact of the dimensionality reduction. 

For version 1 and 2, we have displayed the attractors with dimensions beyond the first three PCA axes to show the dimensionality of these attractors. We can further reason about their intrinsic dimensionality by considering how many components we need to explain 95\% of the variance. For the unscaled version 1 attractor, only 2 PCA components are needed to explain over 95\% of the variance. When scaled, we see that there is interesting behavior in the first 6 dimensions. For the unscaled version 2, 12 components are needed to explain 95\% of the variance. For the scaled version 2, this number jumps to 35 components. And for version 3, unscaled is 17 and scaled is 33. From this, we conclude the attractor manifolds are meaningfully low-dimensional compared to the full dimensionality of the substrate (25600 dimensional). Even so, they will be difficult objects to visualize as they (except for version 1) are too high-dimensional to visualize completely. 

\begin{figure}[h]
    \centering
    \includegraphics[width=0.75\linewidth]{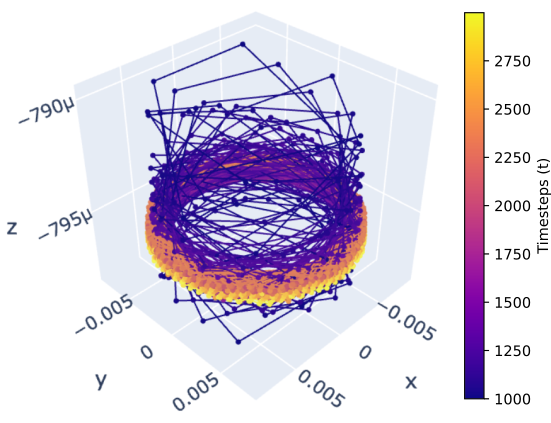}
    \caption{The unscaled version 1 attractor. x,y, and z axes are the first, second, and third most dominant PCA axes. }
    \label{fig:scales_attractor}
    \vspace{-0.5em}
\end{figure}

In fig.\ \ref{fig:scales_attractor}, we show the unscaled version 1 attractor. Here, it becomes clear why only 2 components are needed: The third most dominant PCA-axis, here labeled z, is on the order of $\mu=10^{-6}$. Thus, most movement happens inside the attractor cycle. Note also how small the space of the attractor is: It moves completely inside a circa 0.016$\times$0.017=0.00027 area (unitless).

\paragraph{Lyapunov stability} 

The attractors were run for 4000 steps of burn-in and 10000 steps to estimate the Lyapunov stability. The estimated top 10 Lyapunov values are given in table \ref{tab:lyapunov} (50 for version 3). The $\epsilon$ used was $1*10^{-4}$.


\begin{table}[ht]
    \centering
    \begin{tabular}{l|l|r}
        V & Spectrum, relevant & Simplified \\
        \hline
        1 & $\lambda_1=0.0,$ $\lambda_2=-0.002$, & $0 - \cdots -$ \\
          & $\lambda_{3-5}=-0.007$, $\lambda_{10}=-0.013$ & \\
        \hline
        2 & $\lambda_{1,2}=-0.005$, $\lambda_{3,4}=-0.007$ & $- \cdots -$  \\
          & $\lambda_{5}=-0.01$, $\lambda_{10}=-0.012$ & \\
        \hline
        3 & $\lambda_{1}=-0.009$, $\lambda_{2}=-0.014$ & $- \cdots -$ \\
          & $\lambda_{10}=-0.016$, $\lambda_{50}=-0.031$ &  \\
    \end{tabular}
    \caption{The top Lyapunov exponents of the three attractors. Only informative ones are included.}
    \label{tab:lyapunov}
    \vspace{-1em}
\end{table}

First, the Lyapunov exponents all being negative or 0 tells us that all the attractors are Lyapunov stable. We have no evidence that these systems are chaotic. This means that small perturbations will always shrink, and the system's behavior is likely to be maintained, and easily predicted.

Second, we only have one 0 exponent in version 1, suggesting version 1 has true periodic behavior. For the two others, we have two possibilities: 1) Given the slow convergence of the exponents, there might be 0s that we struggle to estimate. Version 2 then has at least two 0s ($\lambda_{1,2}$ = -0.005), and maybe 4 ($\lambda_{3,4}$ = -0.007), and is therefore likely to be a quasi-periodic attractor. Version 3's spectrum is less clear but might be periodic ($\lambda_1$ = -0.009). Possibility 2) is that both systems could be slowly dissipative, meaning the volume they span is shrinking over time.

\begin{figure}[ht]
    \centering
    \includegraphics[width=0.6\linewidth]{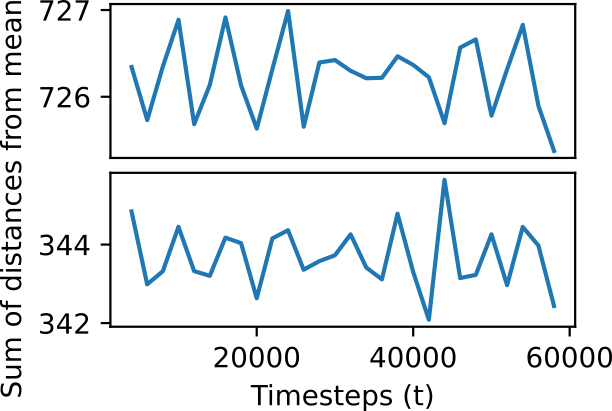}
    \caption{The sum of unscaled distances for version 2 (top) and 3 (bottom). }
    \label{fig:dists}
    \vspace{-0.5em}
\end{figure}

To strengthen either possibility, we ran both systems for 60k timesteps and calculated a proxy for the volume as time progressed. The proxy was the sum of distances from all points to the middle of the attractor. We calculated this every 2000 timesteps. Here, in fig.\ \ref{fig:dists}, we found that both sums oscillate. Version 2's sum has a standard deviation of 0.4 around the mean sum 726, while version 3's sum has an standard deviation of 0.8 around the mean 344. There is no clear downward trend in either sum, suggesting that our failure to find 0s for version 2 and 3 is likely numerical error.

Lastly, we note that we did not calculate all 26000 exponents because it is computationally intensive. As such, we expect there to be more strongly negative exponents in the spectra that were not shown. However, only the top exponent is important for stability, and only the positive and near zero exponents are important for classification, so this does not change our results.

\paragraph{Oscillations and classification} Now, we consider the Fourier power spectra. The attractors were run for 2000 steps of burn-in and 8000 steps to create the spectra. 

\begin{figure}[h]
    \centering
    \includegraphics[width=\linewidth]{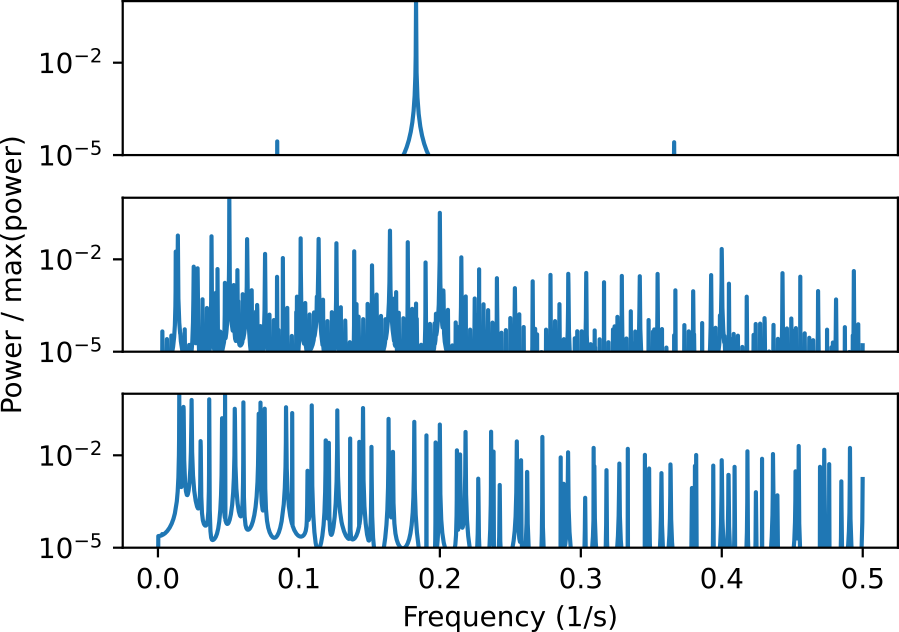}
    \caption{The Fourier power spectra of unscaled data. Top: version 1. Middle: version 2. Bottom: version 3.}
    \label{fig:fourier}
    \vspace{-0.5em}
\end{figure}

Considering the Fourier spectrum in fig.\ \ref{fig:fourier}, version 1 has one clear peak, and a couple smaller ones (scaled power is less than $10^{-4}$). The behavior is dominated by one clear frequency: $f_1$ = 0.183. Together with the Lyapunov spectrum, this clearly indicates true periodic behavior.

Version 2 has a messier Fourier spectrum. There are 23 frequencies above 0.01 in fig.\ \ref{fig:fourier}, and so we will not recount them all here. Many of the peaks are clearly harmonics, and after filtering them out, we are left with 12 frequencies. Filtering away any frequencies that can be described as a linear combination of other frequencies, we are left with 4 frequencies: 0.0126, 0.0141, 0.0422, and  0.0760. We are likely looking at quasi-periodic behavior, as the spectrum suggests 4 dominant frequencies, meaning the quasi-periodic motion is along a 4-torus. This conclusion is also supported by the Lyapunov analysis.    

For version 3, we see that it has 84 frequencies over 0.01.  We note the visual complexity seen in fig.\ \ref{fig:attractor_figure} is clearly corroborated by the Fourier spectra in fig.\ \ref{fig:fourier}. After filtering the frequencies for harmonics and linear combinations, we are left with 3 frequencies: 0.00300, 0.0181, and 0.0476. This is somewhat unexpected, as there was no sharp drop-off in the Lyapunov spectrum after $\lambda_3$. The findings here indicate quasi-periodic motion along a 3-torus.  

\paragraph{Perturbations}

\begin{figure*}[ht]
    \centering
    \includegraphics[width=0.8\linewidth]{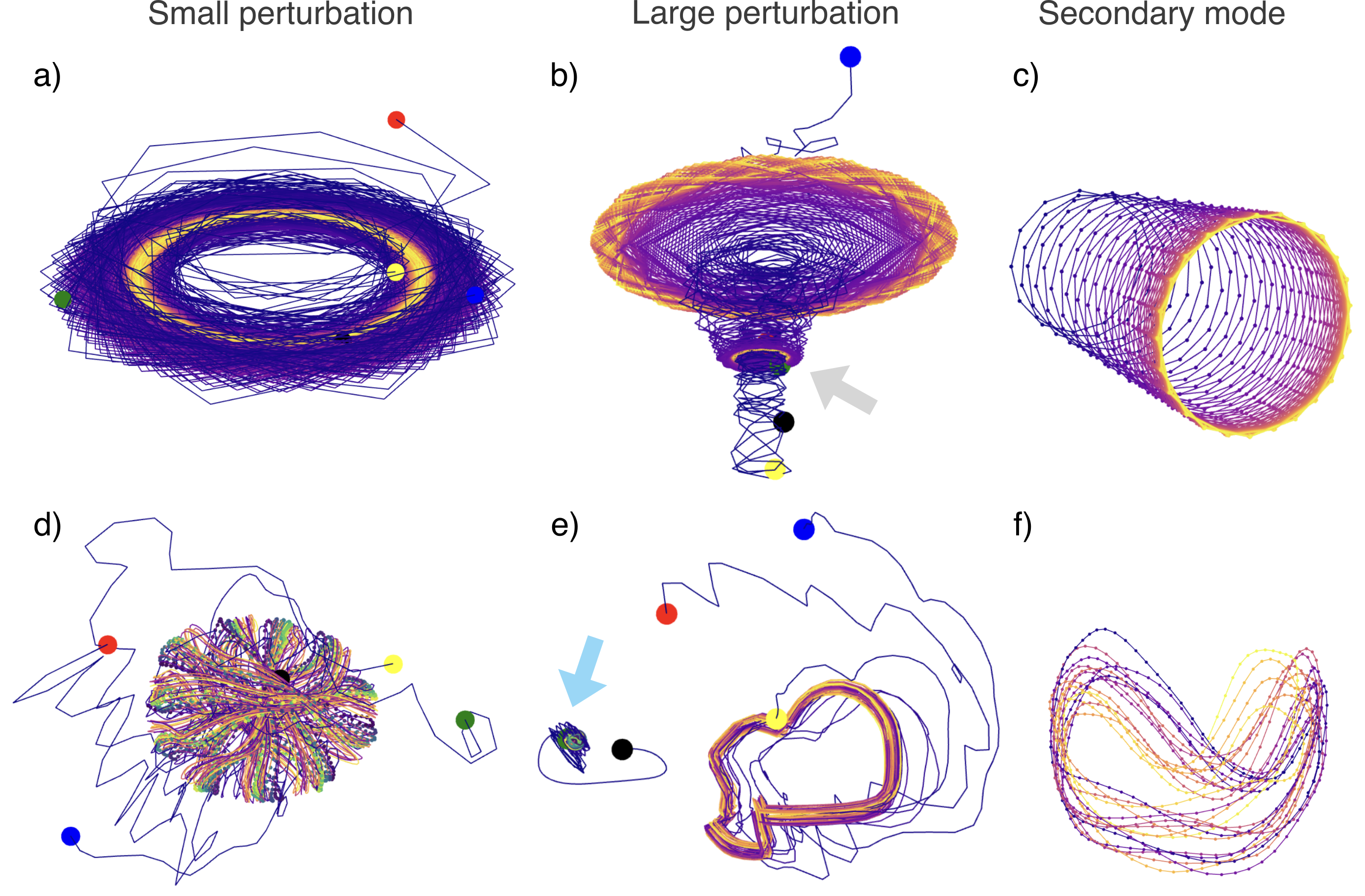}
    \caption{Perturbations returning to system attractors. Top: version 1, unscaled. Bottom: version 3. Time is in colormap ``plasma'', from purple to yellow. Perturbations' start positions have a black, red, green, blue, or yellow ball. Original attractor in ``viridis'' in a, b, d, and e.}
    \label{fig:basins}
    \vspace{-0.5em}
\end{figure*}

We now investigate the basins of attraction. All trajectories after perturbations returned to the goal state. 

First, a small perturbation was applied. In fig.\ \ref{fig:basins}a and d, we see version 1 and 3 respectively recovering from 5 perturbations. All return to the respective attractors and resume the oscillatory motion. 

In fig.\ \ref{fig:basins}b and e, the large perturbation has been applied. We see something unexpected: both systems return to a secondary mode. Version 1 now has what appears to be two limit cycles. The original is the smaller (gray arrow). One of the perturbations that ended up in the new cycle is visualized in fig.\ \ref{fig:basins}c, showing similar behavior to the system before convergence. This cycle's mean position is 0.6 from the original (unscaled, unitless). For comparison, a perturbation that ended up in the original cycle has a mean distance of 0.0009 to the original attractor.

Likewise, for version 3, the secondary mode is a mean distance of 1.7 (unscaled, unitless) away from the original (blue arrow), while the perturbation that ended up back at the original is 0.004 away from the original's mean position. The perturbations that ended up in the secondary mode are also 0.004 away from each other. Version 3's secondary mode has a different geometry from the original attractor, appearing rather as a wavy limit cycle (fig.\ \ref{fig:basins}f). Further, the perturbations that came back to the original attractor came back to its original geometry too.   

\paragraph{Learning}

\begin{figure*}[ht]
    \centering
    \includegraphics[width=\linewidth]{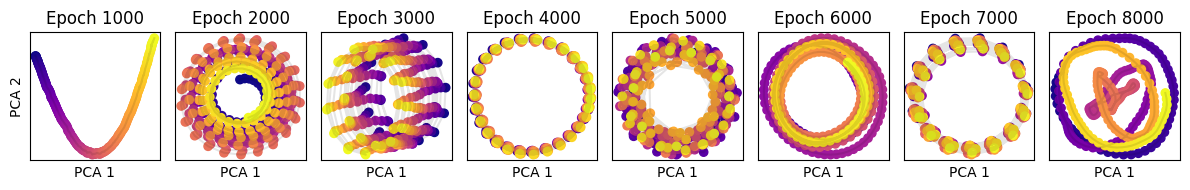}
    \caption{Attractor for model version 3 for 1 to 8*1000 epochs. The plots include 300 timesteps. The colormap is ``plasma'', indicating the timestep number, from purple to yellow.}
    \label{fig:learning}
    \vspace{-1em}
\end{figure*}

As the NCA trained, the models at epoch 1 to 8*1000 were saved. In fig.\ \ref{fig:learning} we show how the attractor forms in version 3 of the green gecko NCA. For the shown training epochs, we ran the NCA until convergence on the attractor (1000 timesteps) and then run for another 300 timesteps while recording from the attractor. Each attractor recording was scaled with standard scaling, and the two dominant principal components have been plotted. 

At epoch 1000, the system has clearly not learned an attractor yet. Up to epoch 2000, some oscillatory attractor has been formed (visually), and $\lambda_1=0.00006$. By epoch 5000 the attractor is skewing dissipative ($\lambda_1=-0.00006$), still close to a limit cycle. However, as seen from the final dynamics in table \ref{tab:lyapunov}, settles on weakly dissipative dynamics instead, with the final $\lambda_1=-0.009$.  

\section{Discussion}

We have used various tools from dynamic systems theory to identify, visualize, and analyze attractors. In addition to what we learned about this case study specifically, this methodology can readily be applied to other NCAs. 

In this case study, we saw that the NCA attractors were not fixed points, and their dynamics were dominantly oscillatory. In fact, oscillatory dynamics were quick to form during learning. Furthermore, oscillations combined with slow convergence means that even though some of these attractors theoretically can converge to a fixed point later, they do not within reasonable time. As such, all relevant dynamics are oscillatory. Oscillatory dynamics have been seen in NCAs before, from its loss \citep{grattarola2021learning}, behavior \citep{randazzo2020self}, and hidden channels \citep{kvalsund_sensormovementdrives, variengien2021towards}. Future work could investigate whether this is universal for NCAs, or simply incidental for these systems.  

Furthermore, we established that perturbations did not return to the attractor properly after perturbations were applied. The rapid convergence along the basin in fig.\ \ref{fig:burn_in} suggests that this secondary mode is not simply the slow dynamics of the basin. Rather, it either entered stable dynamics in the basin, or found a separate attractor altogether. Given that there are infinitely-many possible attractors in the subspace (only constrained by the RGB goal state), we may suggest that perturbations settle on a secondary attractor that is separate from the one it finds from the seed state. There may even be a set of possible attractors that different perturbations converge to. This would mean that the system forms emergent attractors in response to damage. Future work could try to establish the existence of secondary attractors or modes. Additionally, while this may be unique to the non-stochastic gecko NCA, it does posit the question as to whether other NCAs have similar behavior.  

NCAs often have a stochastic update function, but we have here only considered the non-stochastic case. In random dynamical systems, attractors are differently defined \citep{crauel1994attractors}, and the tools used here may need modification to work. Still, we believe that our case study and methodology provide the background for further studies into NCA attractors, stochastic or otherwise.   

When analyzing the dynamics of NCAs, one can do so either at the macro level (substrate) or micro level (cell state) \citep{stovold2026landscapes}. Here we have chosen to do it on the macro level, but it would still be informative to see how the local dynamics affected the global dynamics. It could potentially be better to consider NCAs through the lens of coupled attractors, and the global attractor rather as a synchronization of them.  

Lastly, version 2 and 3 both exhibit dampened, quasi-periodically driven motion, which means that they could be strange non-chaotic attractors (SNAs). `Strange' refers to the geometry having fractal properties. The chaos literature does not tend to mention SNAs (e.g.\ \cite{strogatz2001nonlinear, datseris2022nonlinear}), but has a small, dedicated body of literature. SNAs have been detected in laboratory settings \citep{ditto1990experimental} and in stars \citep{lindner2015strange}, among others. There are methods for testing for them, such as the 0-1 test \citep{gopal2013applicability, bernardini2016overview}, though these can be rather complicated to use well. Still, the possibility is exciting, as SNAs combine properties of order and chaos, something not unfamiliar to cellular automata.   

\section{Acknowledgements}
We would like to thank the rest of the manifold team formed at ALICE, Varun Sharma, Harald Ludwig, and Alexander Mordvintsev, as this work grew from our discussions on the attractor landscape paper. We would also like to thank bioAI's Kosio Beshkov, Leonard Tinius, and Nicolai Haug for their help on dynamical systems theory.

\footnotesize
\bibliographystyle{apalike}
\bibliography{example} 

\end{document}